# PER-DPP Sampling Framework and Its Application in Path Planning

Junzhe Wang

## 1. Abstract

Autonomous navigation in intelligent mobile systems represents a core research focus within artificial intelligence-driven robotics. Contemporary path planning approaches face constraints in dynamic environmental responsiveness and multi-objective task scalability, limiting their capacity to address growing intelligent operation requirements. Decision-centric reinforcement learning frameworks, capitalizing on their unique strengths in adaptive environmental interaction and self-optimization, have gained prominence in advanced control system research.

This investigation introduces methodological improvements to address sample homogeneity challenges in reinforcement learning experience replay mechanisms. By incorporating determinant point processes (DPP) for diversity assessment, we develop a dual-criteria sampling framework with adaptive selection protocols. This approach resolves representation bias in conventional prioritized experience replay (PER) systems while preserving algorithmic interoperability, offering improved decision optimization for dynamic operational scenarios. Key contributions comprise:

Develop a hybrid sampling paradigm (PER-DPP) combining priority sequencing with diversity maximization.Based on this,create an integrated optimization scheme (PER-DPP-Elastic DQN) merging diversity-aware sampling with adaptive step-size regulation. Comparative simulations in 2D navigation scenarios demonstrate that the elastic step-size component temporarily delays initial convergence speed but synergistically enhances final-stage optimization with PER-DPP integration. The synthesized method generates navigation paths with optimized length efficiency and directional stability.

# 2. Preliminaries

## 2.1 Reinforcement Learning[1]

Reinforcement Learning is a learning paradigm where agents autonomously learn to make decision by interacting with an environment, with the goal of maximizing expected rewards. The system is formalized as a Markov Decision Process (MDP), which is defined by a tuple ⟨S,A,P,R,γ⟩, where S represents the state space, A represents the action space, P defines the state transition probabilities, R denotes the reward function, and γ is the discount factor. At each step t, the environment is in a state $s_t$, and the agent selects an action at according to a policy π. The environment then transitions to a new state based on the transition probability $P(s_{t+1} \mid s_t, a_t)$, and the agent receives a reward. The agent's objective is to learn an optimal policy $\pi^*$ that maximizes the expected cumulative discounted reward starting from any initial state $s_t$:

$$V_\pi(s) = E[G_t \mid S_t = s] = E[R_t + \gamma R_{t+1} + \gamma^2 R_{t+2} + .... \mid S_t = s] \quad (2.1)$$

where $V_\pi(s)$ is the value function that estimates the expected return when following policy π from state $s_t$

## 2.2 Experience Replay

In deep reinforcement learning, experience replay plays a critical role by allowing agents to store and reuse historical interactions through a replay buffer. This approach addresses the problem of data correlation inherent in online training processes while enhancing the utilization efficiency of samples. A notable advancement in optimizing experience replay is Prioritized Experience Replay[2] (PER), which strategically weights past experiences to improve learning effectiveness. PER enhances the effectiveness of experience replay mechanisms by selectively emphasizing transitions with higher learning significance, as determined through temporal difference (TD) error measurements, as (2.2).

$$\delta_j = r_j + \gamma \max_a Q(s_{j+1}, a, w_T) - Q(s_j, a_j, w) \quad (2.2)$$

In PER, an experience is assigned a priority $p_j = |\delta_j| + \varepsilon$ where $\varepsilon$ ensures non-zero priority. The probability P(j) of sampling an experience is proportional to its priority:

$$P(j) = \frac{p_j^a}{\sum_k p_k^a} \quad (2.3)$$

By focusing on experiences with higher TD errors, PER enhances learning efficiency and accelerates convergence.

## 2.3 Determinantal Point Processes[3]

There are various methods to measure sample heterogeneity. Many methods struggle to efficiently select highly diverse samples and often require substantial prior knowledge.The Determinantal Point Process (DPP) is a probabilistic model that defines correlations among samples via a kernel matrix, simplifying probability calculations through determinant computations. Elements in the kernel matrix represent pairwise similarities between samples, and the determinant value reflects the degree of heterogeneity within a subset. When a subset contains overly similar elements, the determinant decreases, thereby reducing the probability of their co-occurrence. DPP excels at modeling the balance between diversity and quality of elements in a set and is widely applied in recommendation systems, text summarization, image retrieval, and similar scenarios. Its core idea lies in measuring subset probabilities via matrix determinants, favoring subsets that are both high-quality and diverse. With its unique mathematical formulation and flexible design, DPP provides an efficient and powerful tool for addressing diversity and correlation challenges. A brief introduction is provided below.

Given a predefined sample set Z and its kernel function K , a probability measure space $(Z, 2^Z, P)$ can be mathematically defined. The definitions of Z and P are as follows: Let the candidate sample set Z = $\{z_1, z_2 ... z_N\}$ contain N samples. The Determinantal Point Process (DPP) transforms complex probability calculations into simplified determinant computations, where the probability of sampling any subset $Y \subseteq Z$ is proportional to the determinant of its corresponding kernel submatrix $K_Y$, as shown in Equation (2.4). Here, $K_Y$ denotes the Gaussian kernel matrix associated with the subset Y, which is a submatrix of the original kernel matrix K.

$$P(Y) \propto \det(K_Y) \tag{2.4}$$

The DPP algorithm can be formulated as the following determinant maximization problem: $\arg\max_{Y \subseteq R} \log(\det(K_Y))$ .However, this constitutes an NP-Hard problem. Traditional MAP requires computing determinants over all possible subsets, resulting in exponential complexity (O( $N^3 M^3$ ), where N is the total number of elements and M is the target subset size), which becomes intractable for large-scale datasets. In practical implementations, greedy algorithms[4] are commonly employed to reduce computational complexity to O( $M^2 N$ ) while guaranteeing near-optimal solutions.

The greedy selection process iteratively selects a sample j from the candidate set that maximizes the marginal gain and adds it to the resulting subset Y until a stopping criterion is met, as formalized in Equation (2.5).

$$j = \arg\max_{j \in R \setminus Y} \log\det(K_{Y \cup \{j\}}) - \log\det(K_Y) \tag{2.5}$$

However, due to the high computational complexity of determinant calculations, the Cholesky decomposition of matrices is employed. The specific procedure is as follows: Assume a matrix with its Cholesky decomposition expressed as (2.6), where V is a non-invertible lower triangular matrix, and $\det(K_Y) > 0$

$$K_Y = VV^T \tag{2.6}$$

For any $j \in R \setminus Y$, $K_{Y \cup \{j\}}$ has:

$$\begin{aligned} K_{Y \cup \{j\}} &= \begin{bmatrix} K_Y & K_{Y,j} \\ K_{Y,j}^{\ T} & K_{jj} \end{bmatrix} = \begin{bmatrix} V & 0 \\ C_j & d_j \end{bmatrix} \begin{bmatrix} V & 0 \\ C_j & d_j \end{bmatrix}^T \\ &= \begin{bmatrix} VV^T & VC_j^{\ T} \\ C_j V^T & C_j C_j^{\ T} - d_j^{\ 2} \end{bmatrix} \end{aligned} \tag{2.7}$$

Then we get:

$$VC_j^{\ T} = K_{Y,j} \tag{2.8}$$

$$d_j^{\ 2} = K_{jj} - \left\| C_j \right\|_2^2 \tag{2.9}$$

$$\det(K_{Y \cup \{j\}}) = \det(VV^T) \bullet d_j^{\ 2} \tag{2.10}$$

According to (2.10), we can simplify (2.5) as follows:

$$i = \arg\max_{j \in R \setminus Y} \log(d_j^{\ 2}) \tag{2.11}$$

The advantage of this method lies in transforming the Cholesky decomposition process into an incremental computation rather than direct decomposition when adding new samples. After incorporating sample i , obtained through Equation (2.11), into the acquired subset Y , the updated Cholesky decomposition of the sub-kernel matrix can be derived according to Equation (2.7) as follows:

$$K_{Y \cup \{i\}} = \begin{bmatrix} V & 0 \\ C_i & d_i \end{bmatrix} \begin{bmatrix} V & 0 \\ C_i & d_i \end{bmatrix}^T = V' V'^T \tag{2.12}$$

Similar to(2.8), $C_i$ and $d_i$ are updated and recorded, for every $j \in R \setminus (Y \cup \{i\})$ we can get a new decomposition as follows:

$$V' C_j'^{\,T} = \begin{bmatrix} V & 0 \\ C_i & d_i \end{bmatrix} C_j'^{\,T} = K_{Y \cup \{i\}, j} = \begin{bmatrix} K_{Y,j} \\ K_{ij} \end{bmatrix} \tag{2.13}$$

$$C_j'^{\,T} = \begin{bmatrix} V & 0 \\ C_i & d_i \end{bmatrix}^{-1} \begin{bmatrix} K_{Y,j} \\ K_{ij} \end{bmatrix} = \begin{bmatrix} V^{-1} K_{Y,j} \\ -(C_i V^{-1} K_{Y,j} - K_{ij}) / d_i \end{bmatrix} \tag{2.14}$$

Then with (2.8) and (2.14),we can get:

$$C_j' = \begin{bmatrix} C_j & (K_{ij} - C_i^T C_j) / d_i) \end{bmatrix} \triangleq [C_j \quad e_j] \tag{2.15}$$

$$d_j' = K_{jj} - \left\| C_j' \right\|_2^2 = d_j^2 - e_j^2 \tag{2.16}$$

## 2.4 Elastic DQN[5]

The Elastic DQN algorithm primarily integrates the concepts of Coarse Q-Learning and multi-step DQN learning, leveraging their distinctive properties to mitigate overestimation and enhance the overall performance of DQN. First, to incorporate Coarse Q-Learning principles, a memory bank is introduced before the experience replay buffer. This module employs unsupervised clustering analysis to evaluate the similarity between the current state and previous states. Meanwhile, multi-step DQN exhibits sensitivity to the hyperparameter controlling the number of learning steps. The memory bank dynamically adjusts learning steps by aggregating updates for similar states into a single operation while processing dissimilar states independently, thereby enabling adaptive step-size updates. The algorithm workflow is illustrated in Figure 2.1.

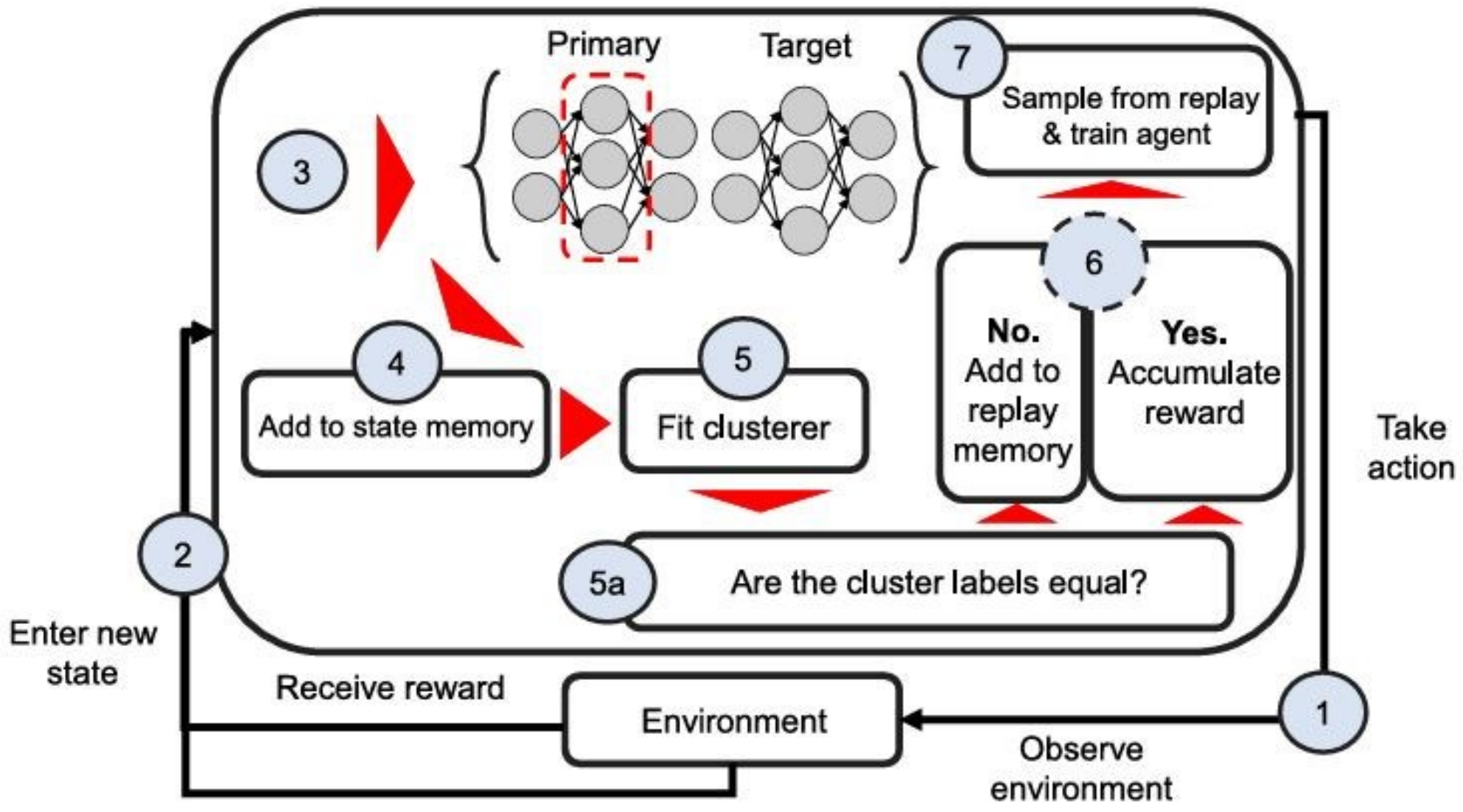

Figure 2.1 Elastic DQN workflow

# 3. PER-DPP-Elastic DQN

## 3.1 PER-DPP sampling paradigm

Fujimoto et al[6] demonstrated that prioritized sampling may excessively focus on a small subset of samples with high temporal-difference (TD) errors, leading to over-reuse of specific samples and consequently reducing sample diversity. Fedus et al[7] further noted that the prioritization mechanism in PER introduces distributional bias, causing models to overemphasize early high-error samples, which may not optimally benefit long-term learning. Li et al[8] improved algorithmic efficiency by filtering high-similarity sequences duringexperience replay to reduce redundancy. Zhao et al[9] proposed incorporating sample diversity into batch sampling, where higher heterogeneity among samples accelerates agent learning. To address PER-induced diversity reduction caused by overemphasis on high-TD-error samples, this study introduces a two-stage hybrid algorithm. The first stage employs PER for importance calculation and ranking to select a larger batch of experiences, followed by the Fast Greedy MAP algorithm to extract a subset with enhanced diversity from this batch.

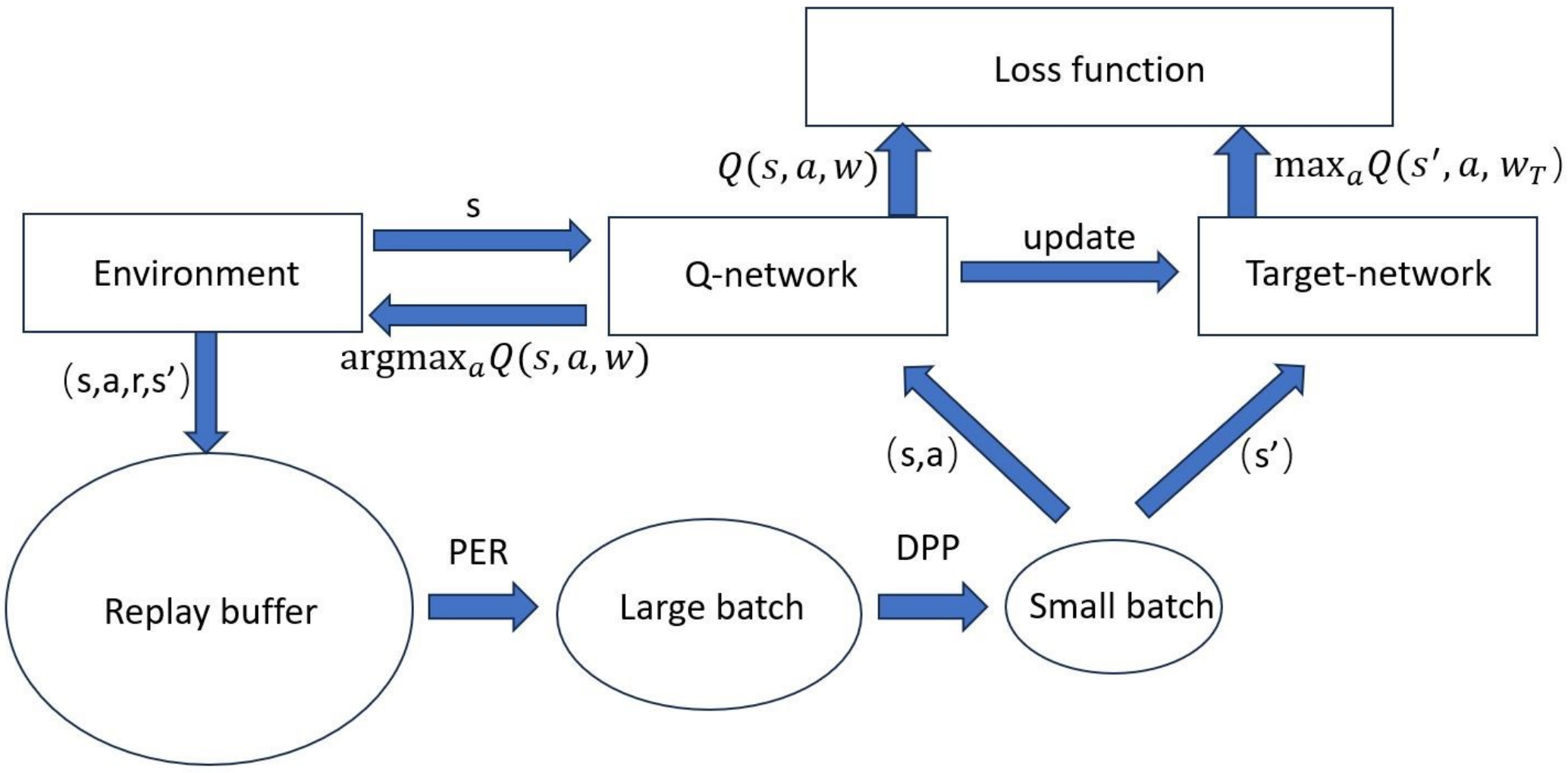

Figure 3.1 PER-DPP workflow

## 3.2 PER-DPP-Elastic DQN Algorithm

Similar to (2.2), we found that for multi-step DQN, TD error is as follow:

$$\delta_j = \sum_{k=0}^{n-1} \gamma^k r_{t+k+1} + \gamma^n \max_a Q(s_{j+n}, a, w_T) - Q(s_j, a_j, w) \qquad (3.1)$$

The Elastic DQN algorithm stores step-count information in the experience replay buffer and utilizes a multi-step DQN approach for agent network parameter updates. When integrated with the PER-DPP sampling framework, corresponding modifications to priority calculations are required. The associated pseudocode is provided as the following Table 3.1:

Table 3.1 Pseudo code for PER-DPP- Elastic DQN

**Initialization**：step length d=0, replay buffer D, memory bank B, Setting the target network and main network with the same shape and initial parameters.

While not finished：

For every time step t：

**Experience sample clustering judgment or storage:**

1.with $\varepsilon$ -greedy policy, get action $a_t$ from $s_t$

2.get next state $s_{t+d+1}$ and reward $r_t$ ,compute the Q value of $s_t$ and $s_{t+d+1}$

3.store $Q(s_t)$ and $Q(s_{t+d+1})$ into memory bank B

4.get samples from B to apply clustering,and add Q value into the samples

5. Using the HDBSCAN, make the following judgments on the results：

If $Q(s_t)$ and $Q(s_{t+d+1})$ have the same lable：

store $(s_{t,}a_t, R_t, s_{t+d+1}, d)$ into D，with most priority $p_t$

reset d=0

else,compute accumulative reward：

$R_t += \gamma^d r_{t+d+1}$， d+=1

**Sample and update network parameters:**

6.sample $j \sim P(j) = \frac{p_j^a}{\Sigma_k p_k^a}$

7.compute weight $w_j = (N \cdot P(j))^{-\beta} / \max_k w_k$

8.update the priority $p_j = |\delta_j| = |R_j + \gamma^{d+1} \max_a Q(s_{j+d+1}, a, w_T) - Q(s_j, a_j, w)|$

9.calculate the kernel matrix for measuring sample similarity

10. Using the Fast Greedy MAP to select small batches of empirical samples

11. Update network parameters based on experience and weights

$s_t \leftarrow s_{t+1}$

**Copy the main network parameters to the target network every T times**

## 3.3 Experimental design

This section briefly introduces the two-dimensional maze environment for path planning (as shown in Figure 3.2), including the design of state space, action space, and reward function, and presents the experimental results of PER-DPP-Elastic DQN on three maps in the above environment.

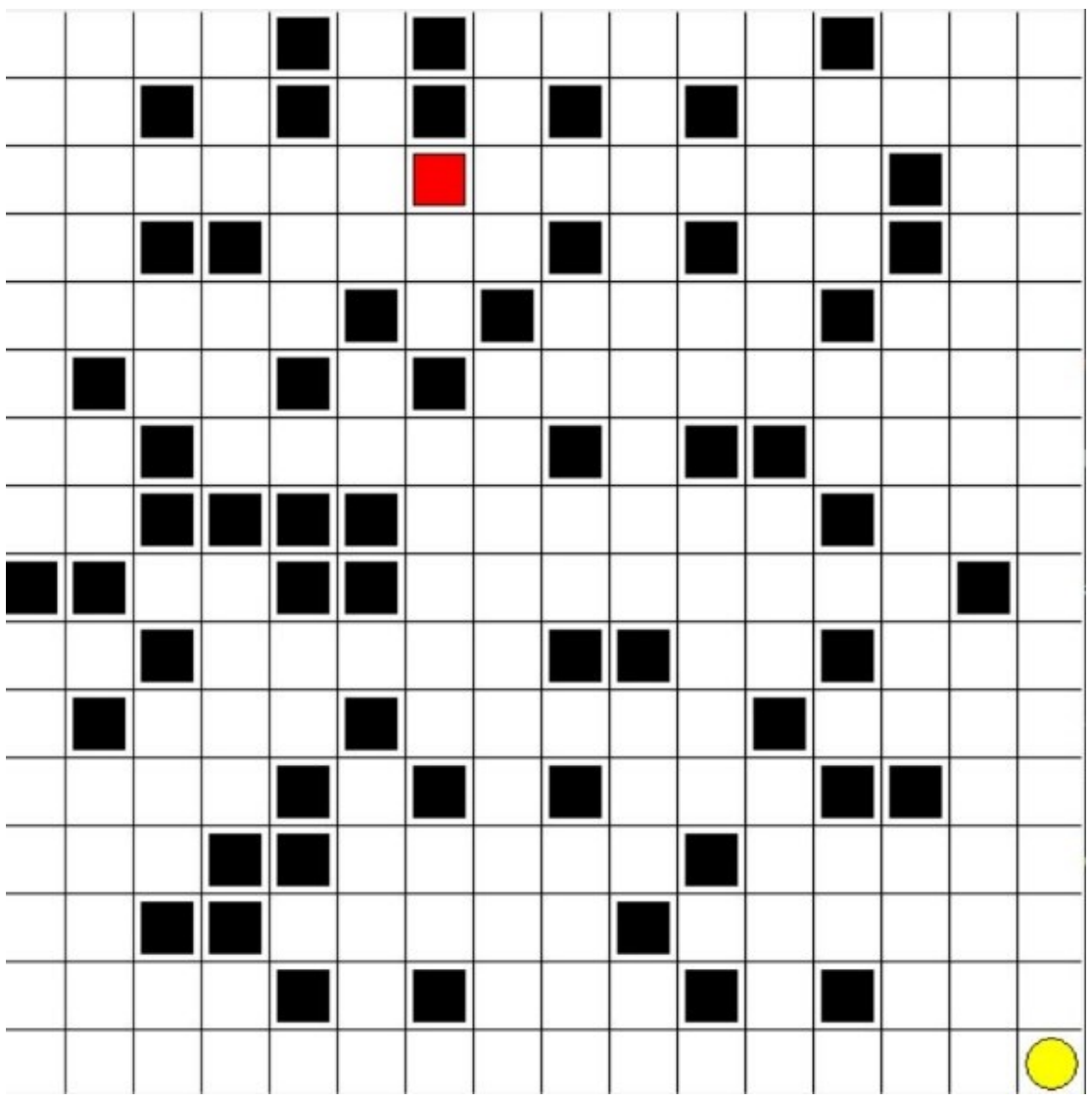

Figure 3.2 Two dimensional maze environment

in the exploration task of unknown environments, this study constructed three core state parameters as shown in Table 3.2:

Table 3.2 State Space for Two Dimensional Maze Environment

| Parameter | Parameter meaning |
|---|---|
| dx | horizontal distance from the target point |
| dy | vertical distance from the target point |
| ob | information on 8 nearby obstacles |

Based on the actual situation, this study sets up an 8-dimensional action space, representing 8 directions on a two-dimensional plane: front, back, left, right, and left front, left back, right front, right back. The executed action moves one unit distance in the corresponding direction, as shown in Figure 3.3:

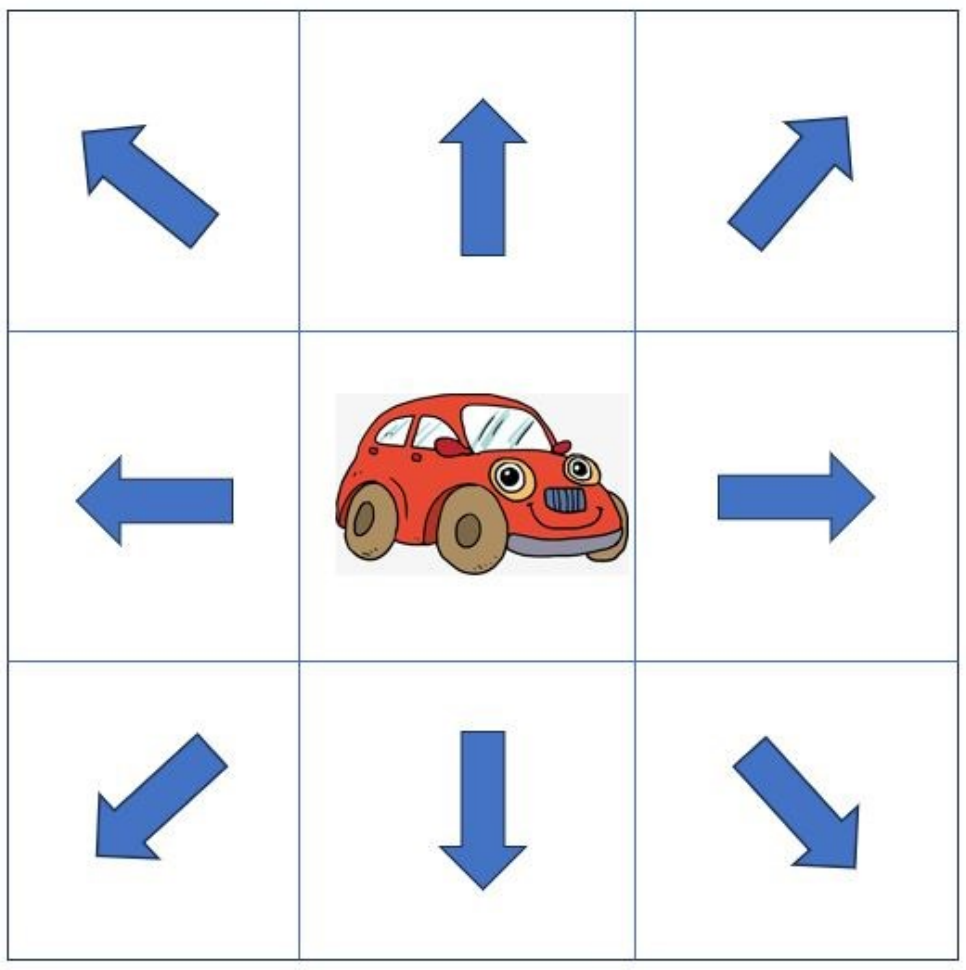

Figure 3.3 maze environment action space

The design of the reward function r is shown in (3.2), which is relatively simple and easy to understand. If the agent remains stationary, a penalty of -200 will be given to encourage the agent to explore the environment; If the intelligent agent reaches its destination after performing an action, a large reward of 500 will be given; If the intelligent agent encounters an obstacle after performing an action, a large punishment of -500 will be given; If the intelligent agent is closer to the target after performing an action, a reward of 100 will be given; If the target point is far away, a penalty of -100 will be imposed.

$$r = \begin{cases} -500 & \text{encounter obstacles} \\ -200 & \textit{remain stationary} \\ -100 & \text{move further away from the target} \\ 100 & \text{get closer to the target} \\ 500 & \text{reaching the target} \end{cases} \quad (3.2)$$

## 3.4 Experimental result presentation

The experiment utilized the Tkinter library on the VSCode platform to create a 16 * 16 simulation grid environment, as shown in Figure 3.4. The white cells in the environment represent accessible areas, red squares represent agents, yellow ellipses represent destinations, and black squares represent obstacles. In order to ensure the universality of the algorithm under different difficulty levels, three maps with different characteristics of obstacle layouts were designed.

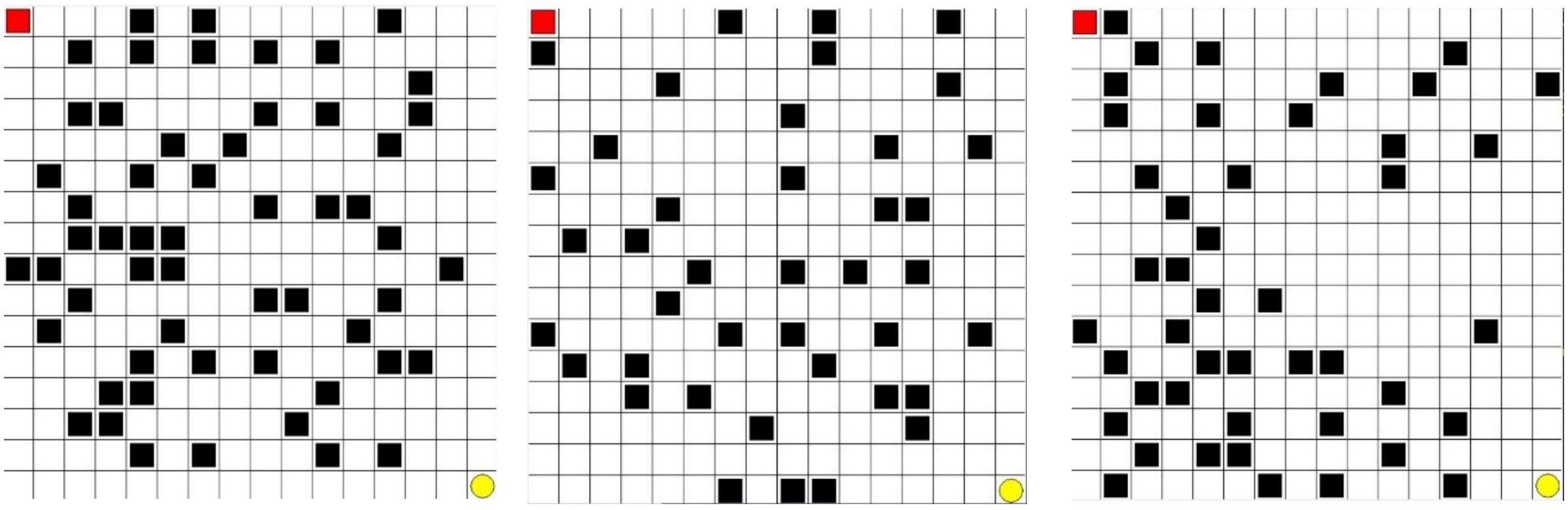

Figure 3.4 three maps

Among them, the obstacles in Map 1 are random distributed and have a high density; Obstacles are random distributed and relatively sparse in Map 2 ; The obstacle design in Map 3 has a certain degree of guidance, with obstacles concentrated in the lower left corner. In the early stages of exploration, there is only one feasible path that approaches the lower left corner. The agent needs to learn the path to transfer to the lower right corner in the middle and later stages in order to successfully reach the target. The presentation and analysis of the training results are as follows:

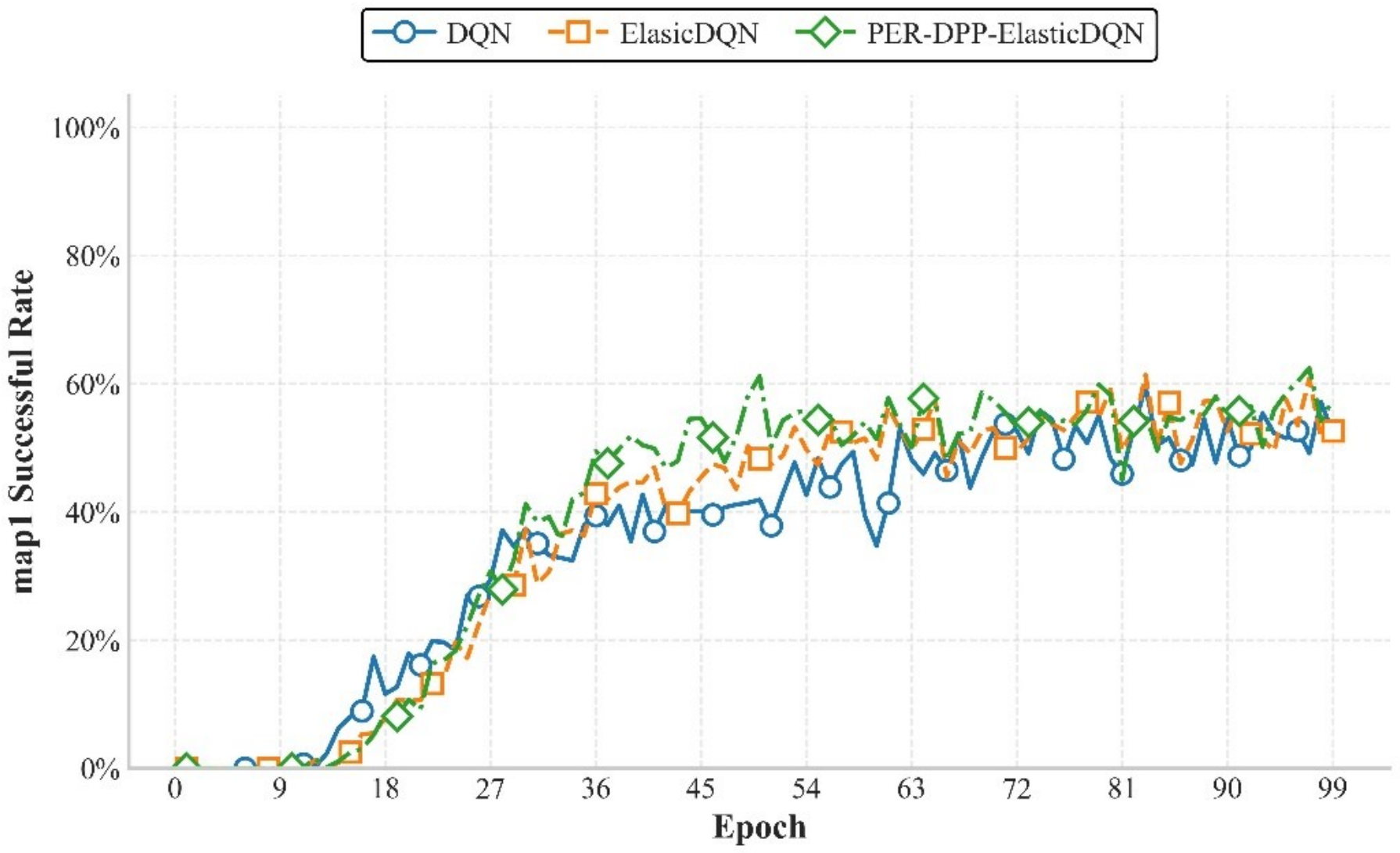

Figure 3.5 Successful Rate Convergence Curve of Map 1

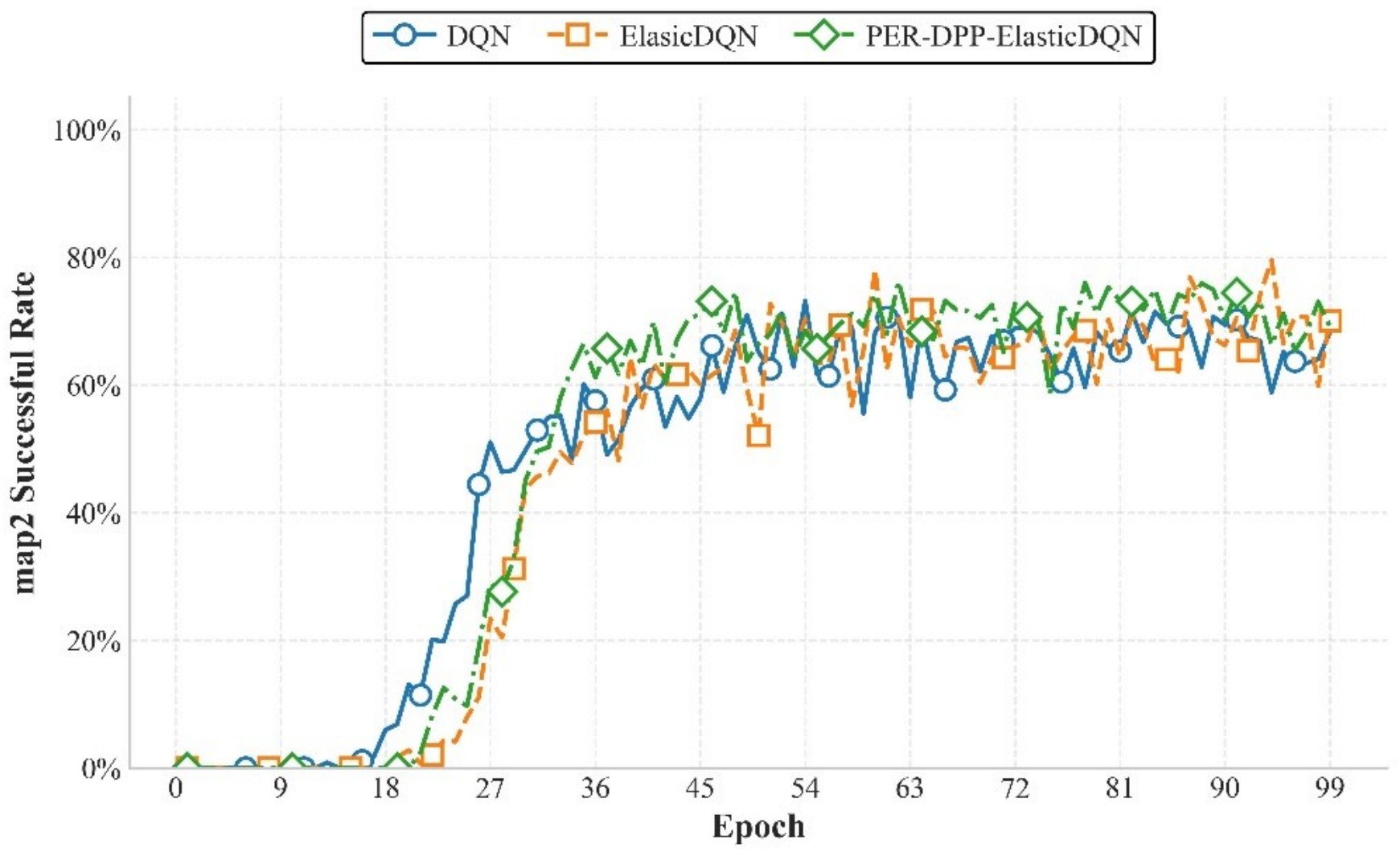

Figure 3.6 Successful Rate Convergence Curve of Map 2

In the early stages of training, the convergence curves of the successful rate within the epoch of Map 1 and Map 2 are similar (where the successful rate within the epoch refers to the average successful rate of all complete rounds in the current epoch), as shown in Figure 3.5 and 3.6. During the experiment, it was observed that the elastic step mechanism of Elastic DQN adopted a large average number of steps in the early stages of training, and the data update of the experience pool was relatively slow. The speed of model training successful rate increase was not as fast as that of DQN. However, as the training data collected from the experience pool gradually increased, the number of elastic steps decreased, and the training process accelerated

We define the average successful rate of the agent during the last 10 epochs as the final convergence successful rate of the algorithm. On Map 1, the final convergence successful rate of standard DQN is 51.9%, and the curve reaches for the first time in the 63rd epoch; The final convergence successful rate of Elastic DQN is 54.1%, and the curve reaches for the first time in the 62nd epoch; The final convergence successful rate of PER-DPP ElasticDQN is 56.7%, and the curve reaches for the first time in the 50th epoch. On Map 2, the final convergence successful rate of standard DQN is 66.2%, and the curve reaches for the first time in the 47th epoch; The final convergence successful rate of Elastic DQN is 69.1%, and the curve reaches for the first time in the 53rd epoch; The final convergence successful rate of PER-DPP ElasticDQN is 70.6%, and the curve reaches for the first time in the 46th epoch. The above results indicate that introducing the PER-DPP sampling framework based on the Elastic DQN algorithm can accelerate the convergence of the model to a certain extent and improve the successful rate of path planning.

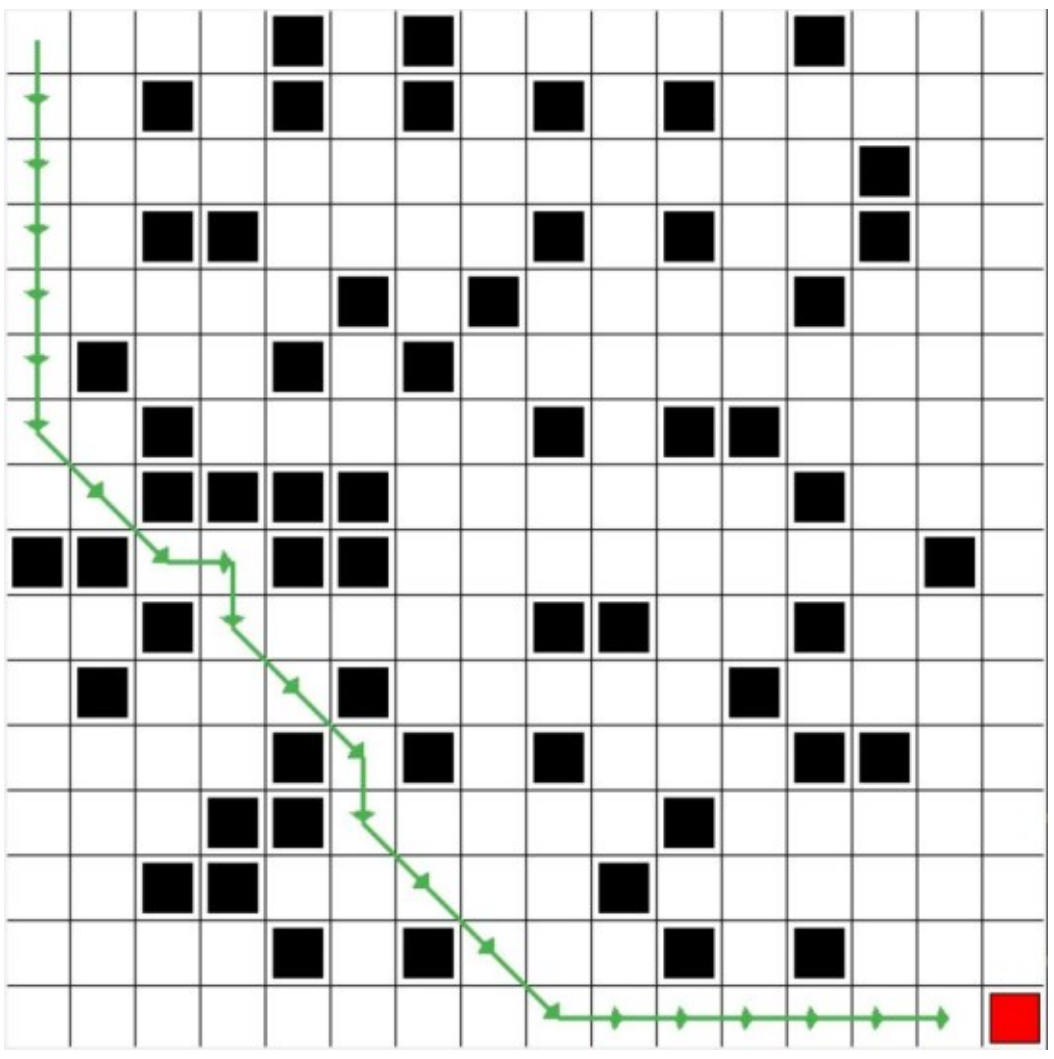

Figure 3.7(a) DQN path in map1

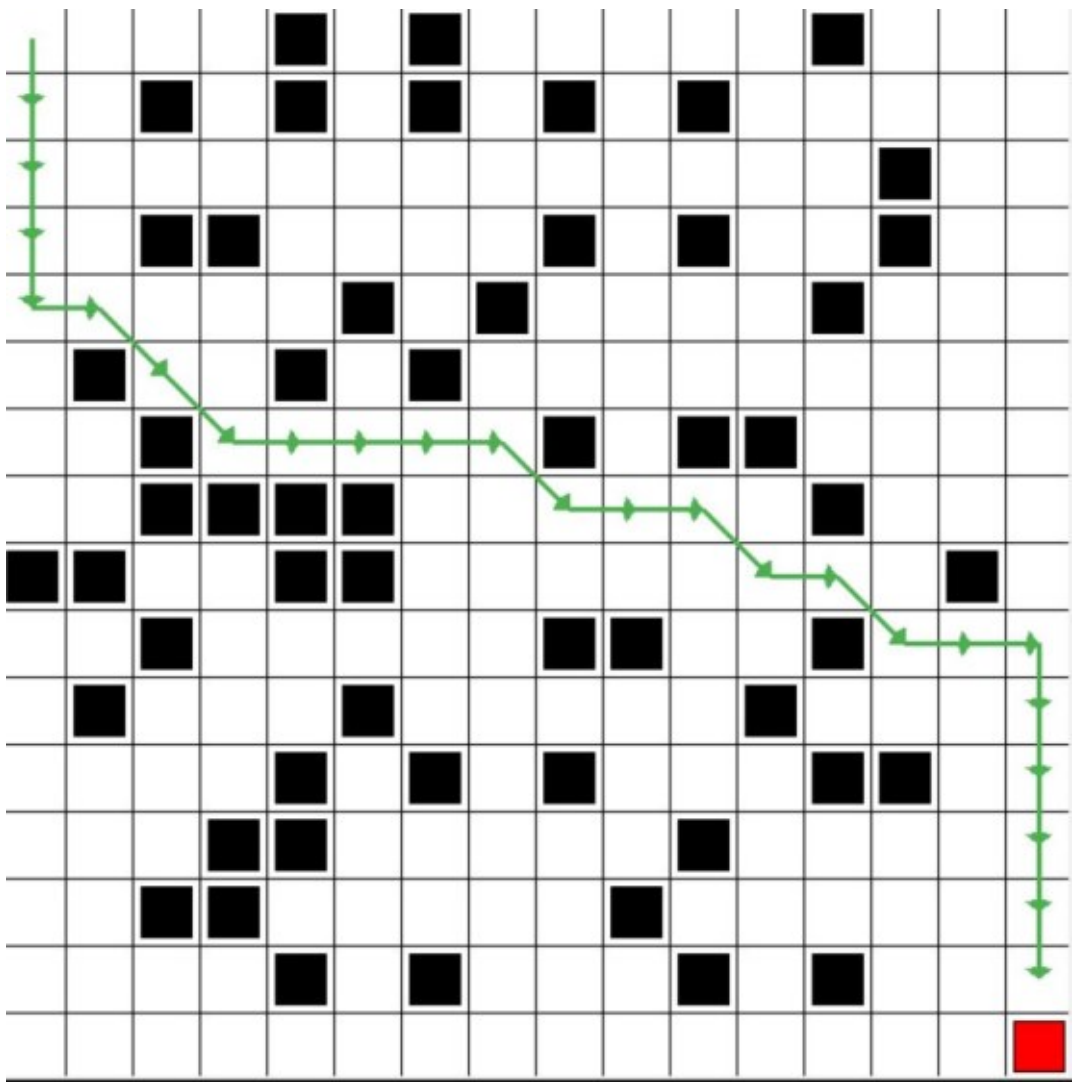

Figure 3.7(b) Elastic DQN path in map1

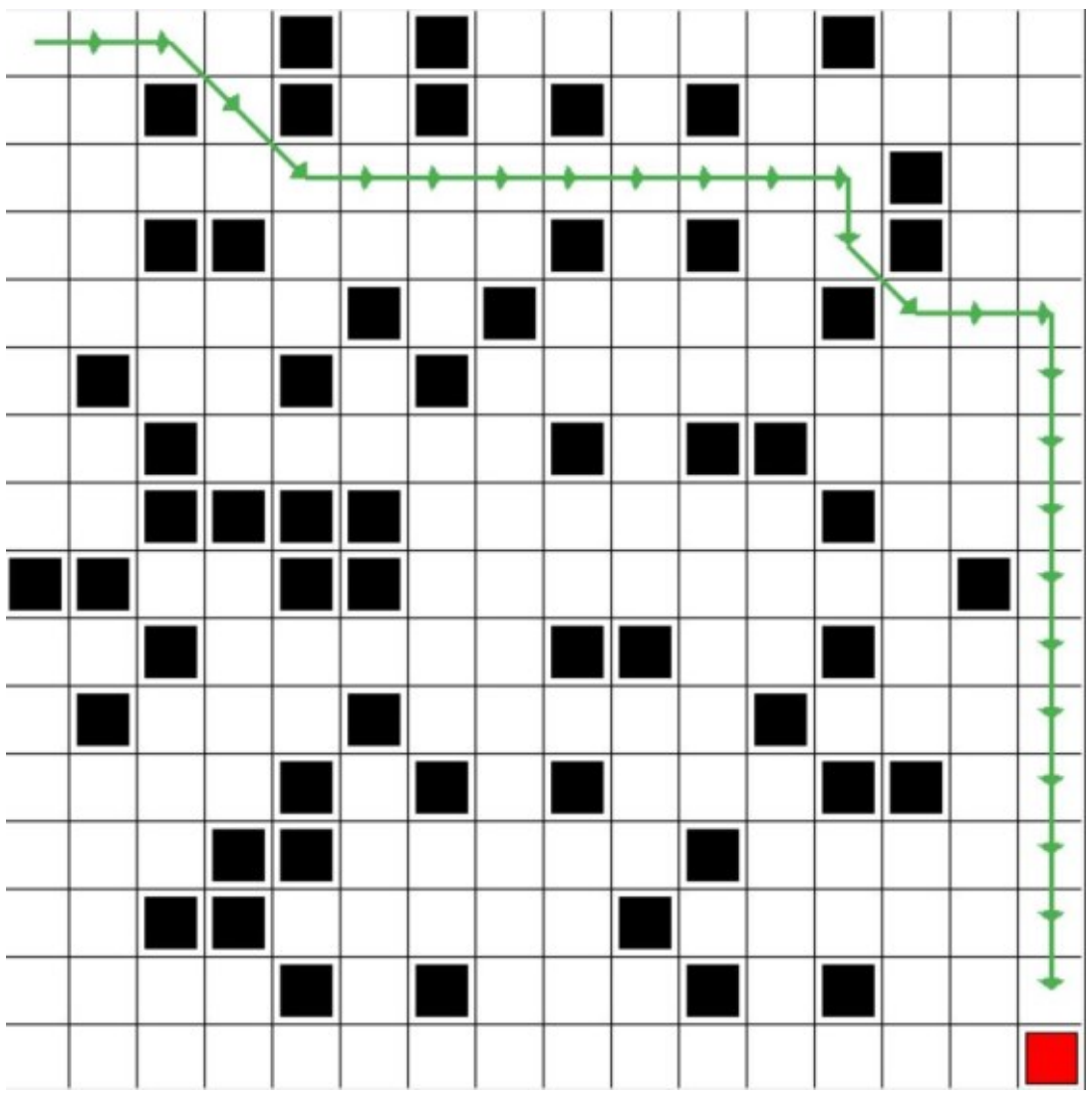

Figure 3.7(c) PDE path in map1

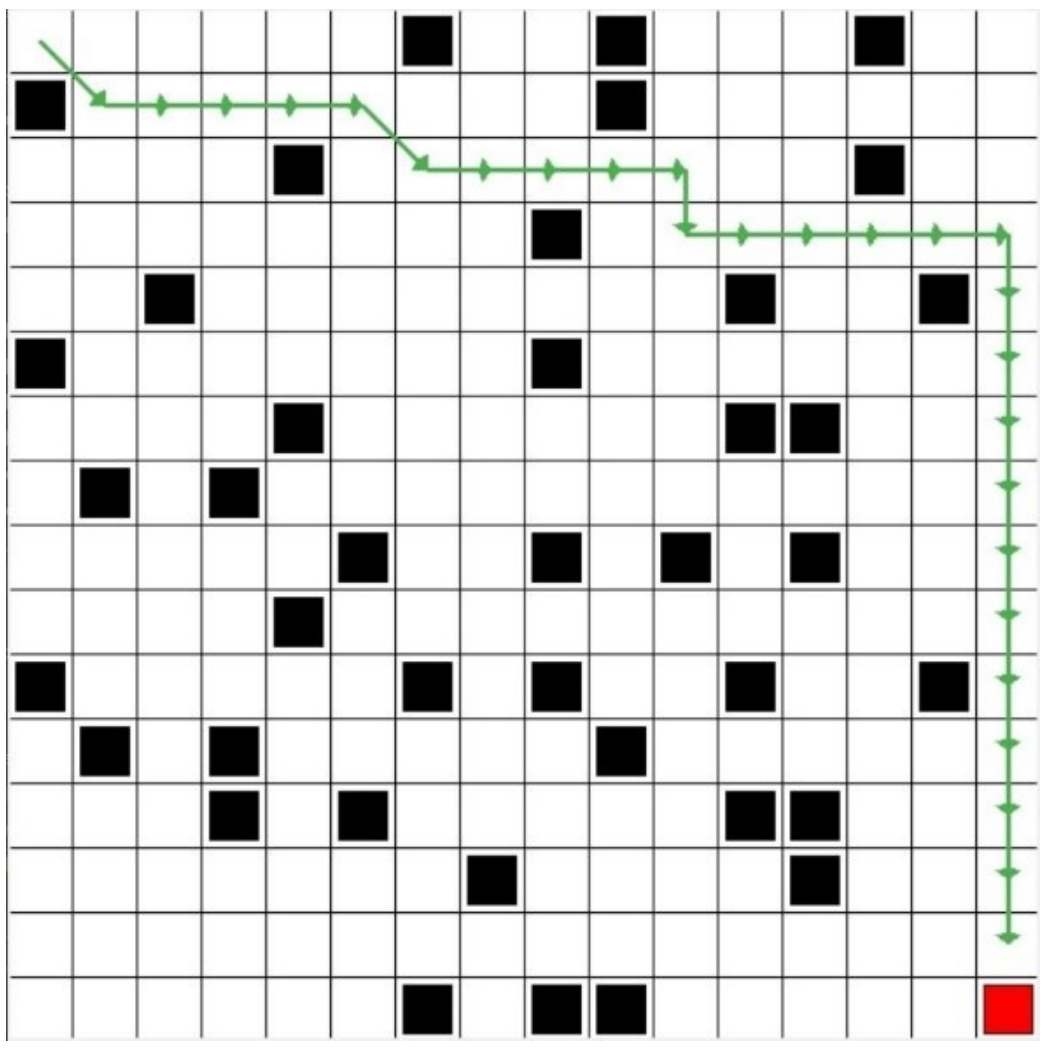

Figure 3.7(d) DQN path in map2

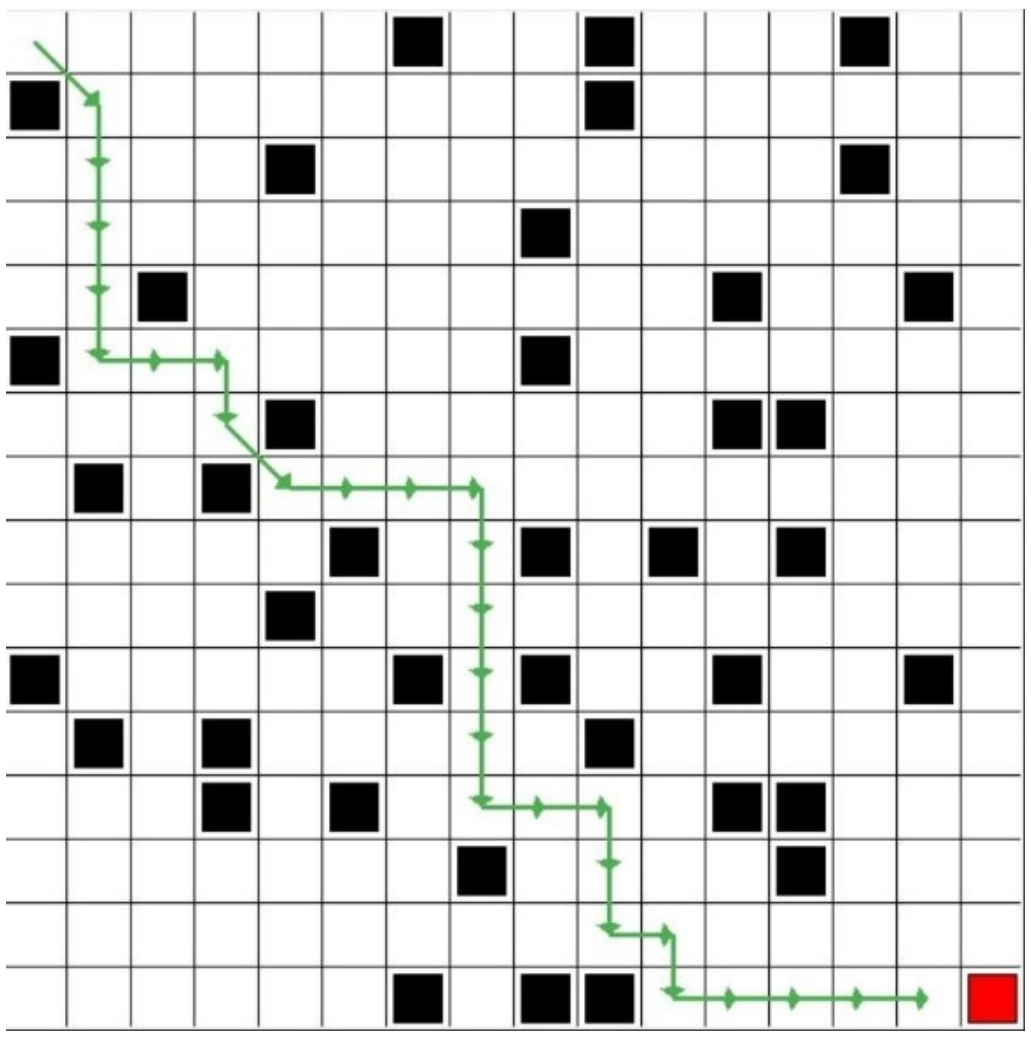

Figure 3.7(e) Elastic DQN path in map2

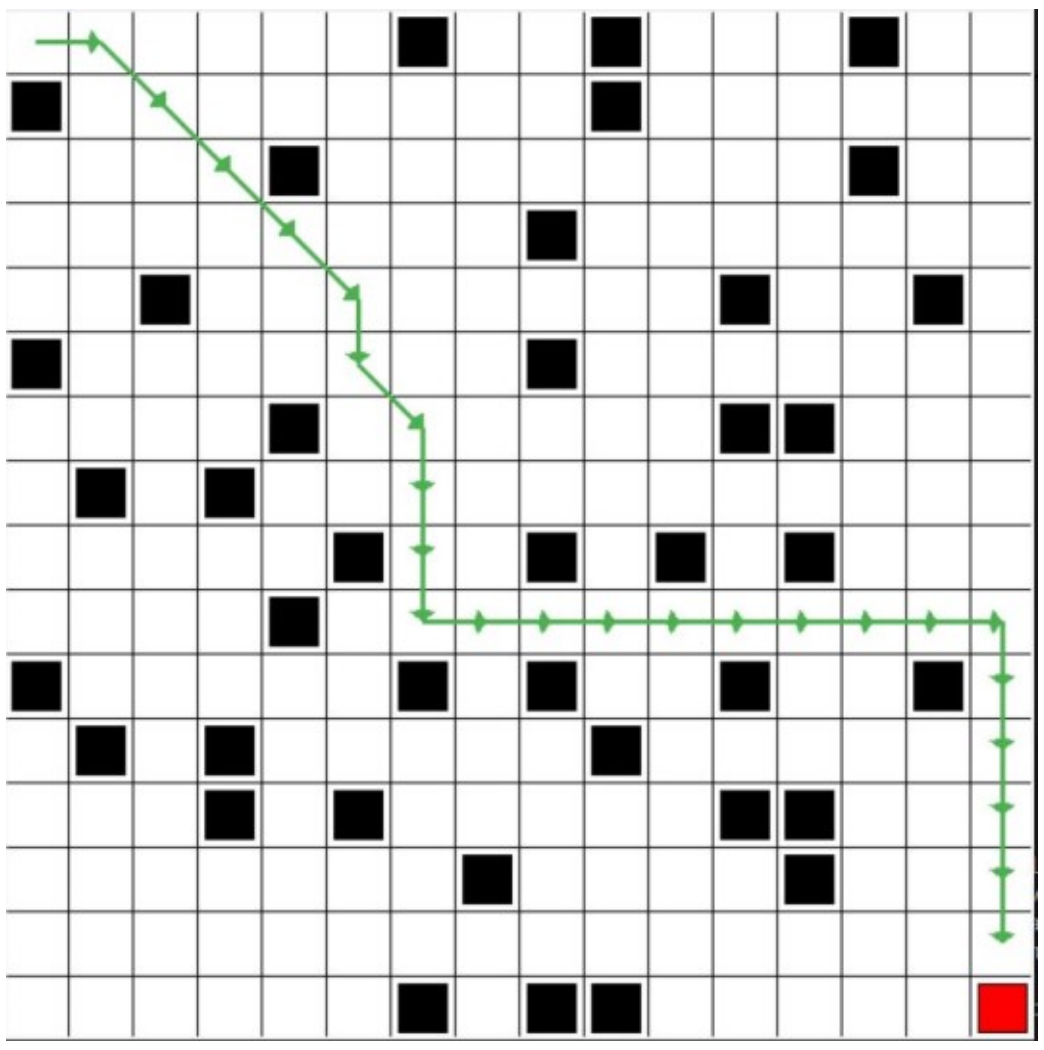

Figure 3.7(f) PDED path in map2

The optimal path, path length, and number of turns planned by three algorithms on two maps are shown in Figure 3.7 and Table 3.3(PER-DPP-Elastic DQN is abbreviated as PDED)

Table 3.3 Three algorithms for optimal path information on Map 1 and Map 2

| Algorithm | Map | Length | Number of turns |
|---|---|---|---|
| DQN | Map1 | 27 | 6 |
| Elastic-DQN | Map1 | 25 | 10 |
| PER-DPP-ElasticDQN | Map1 | 23 | 7 |
| DQN | Map2 | 28 | 6 |
| Elastic-DQN | Map2 | 28 | 11 |
| PER-DPP-ElasticDQN | Map2 | 25 | 6 |

On Map 1, although the optimal path length of Elastic DQN has been reduced compared to standard DQN, the number of turns has significantly increased; There is also a similar trend in the optimal path turning times between Elastic DQN and DQN on Map 2. After further introducing the PER-DPP sampling framework, Map 1 and Map 2 showed better performance in terms of optimal path length and number of turns.

Compared to Map 1 and Map 2, the obstacle distribution in Map 3 is more unique, with the average successful rate curve and optimal path shown in Figures 3.8 and 3.9. The optimal path length and number of turns are shown in Table 3.4

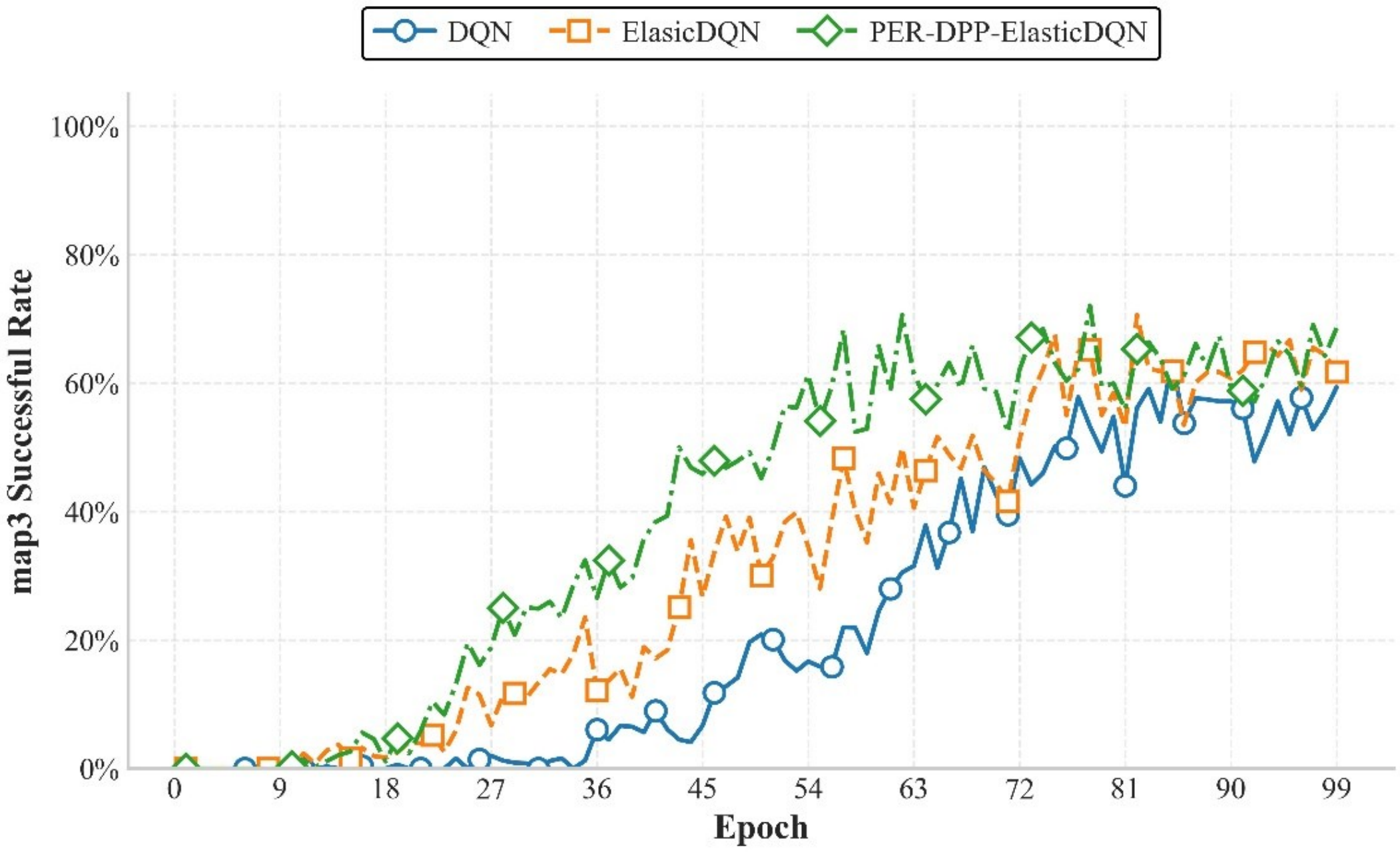

Figure 3.8 Successful Rate Convergence Curve of Map 3

On Map 3, the final convergence successful rate of DQN is 55%, and the curve reaches the final convergence success rate for the first time in epoch 78; The final convergence successful rate of Elastic DQN is 64.3%, and the curve reaches the final convergence success rate for the first time in the 76th epoch; The final convergence success rate of PER-DPP-ElasticDQN is 64.2%, and the curve reaches the final convergence successful rate for the first time in the 58th epoch

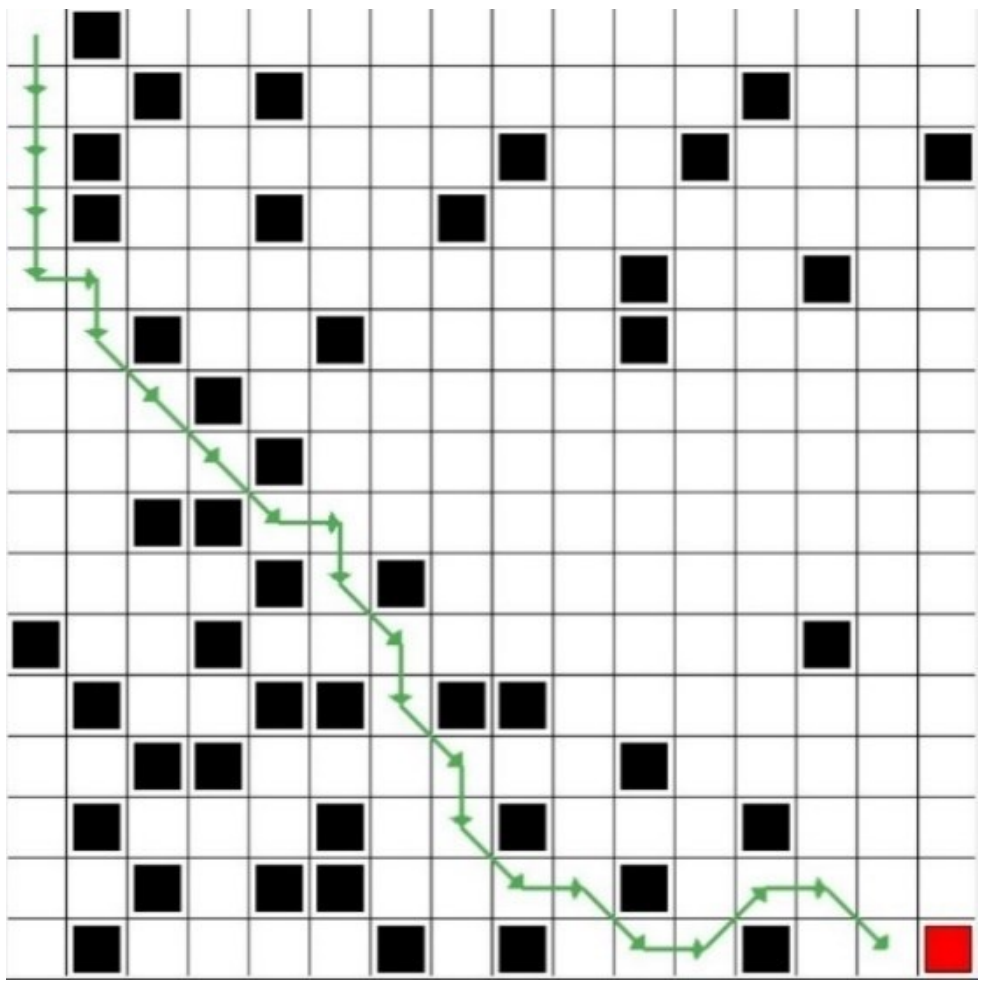

Figure 3.9(a) DQN path in map3

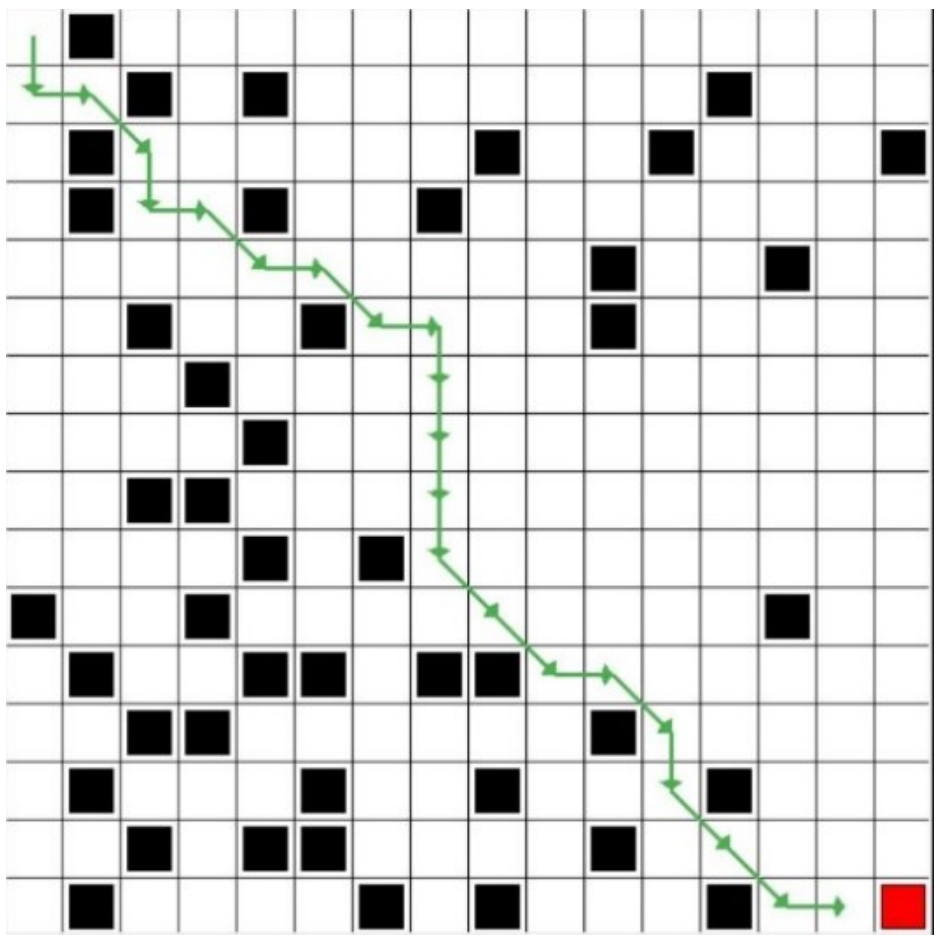

Figure 3.9(b) Elastic DQN path in map3

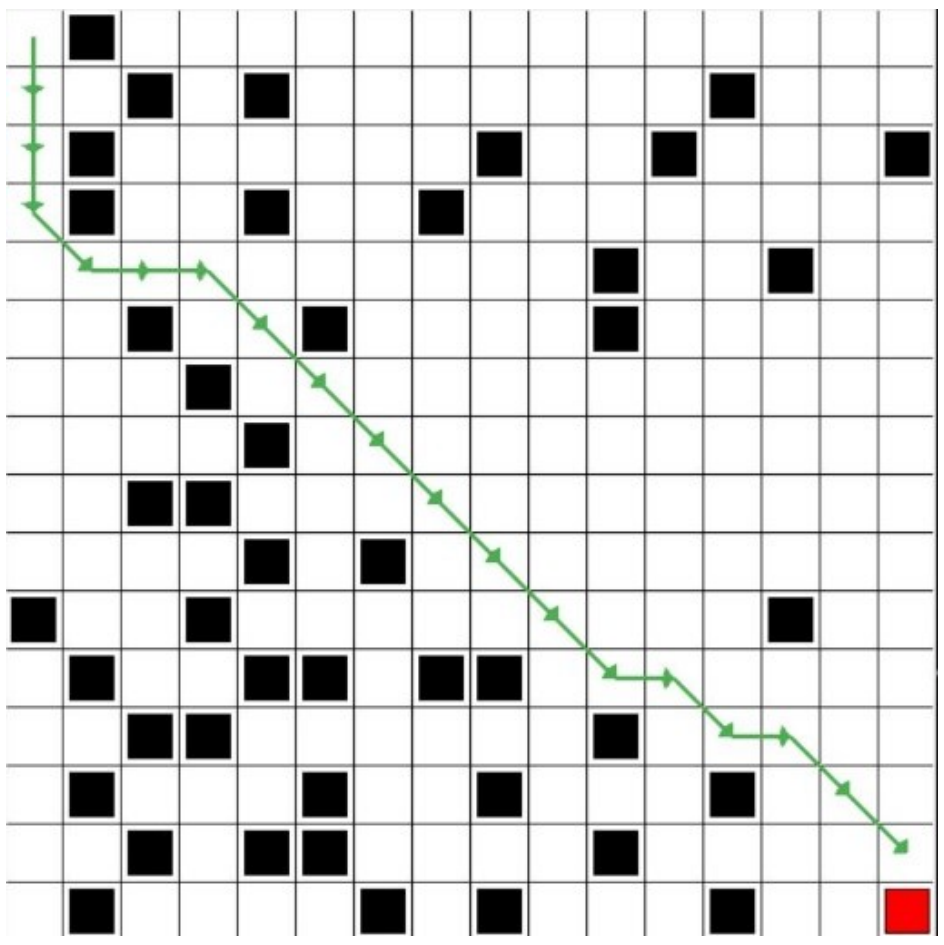

Figure 3.9(c) PDED path in map3

Table 3.4 Three algorithms for optimal path information on Map 3

| Algorithm | Map | Length | Number of turns |
|---|---|---|---|
| DQN | Map3 | 23 | 16 |
| Elastic-DQN | Map3 | 22 | 15 |
| PER-DPP-ElasticDQN | Map3 | 19 | 8 |

The optimal path lengths of the three algorithms on Map 3 decrease sequentially. The PER-DPP-ElasticDQN algorithm also significantly reduces the number of turns required for the optimal path compared to the other two algorithms. It is worth noting that unlike Map 1 and Map 2, the DQN algorithm did not show an early increase in success rate during the initial training stage. Observing the experimental process, it was found that in the early training stage, the agent tended to continuously take downward actions and enter the area with dense obstacles in the lower left corner. Therefore, the convergence curve of the success rate only showed signs of gradually increasing after about 36 epochs. This may be because the PER-DPP mechanism can help the agent learn rich empirical information earlier.

This chapter combines the PER-DPP sampling framework with the Elastic DQN algorithm to form the PER-DPP-Elastic DQN algorithm. Three maps with different characteristics were designed in a two-dimensional maze environment, and the training results of DQN, Elastic DQN, and PER-DPP Elastic DQN were compared on them. In Map 1 and Map 2, introducing the Elastic step mechanism during the initial training stage can result in a higher number of learning steps and a slower learning process compared to DQN. However, with the accumulation of empirical data, the PER-DPP-Elastic DQN algorithm can help agents learn paths with better performance in both path length and turning times at a faster speed. In addition, the experimental results in Map 3 indicate that PER-DPP ElasticDQN is more adaptable to environments with special information compared to DQN.